%% file: 0_main.tex
\documentclass[11pt,letterpaper,logo,twocolumn]{style}

\usepackage[numbers]{natbib}
\usepackage{graphicx}
\usepackage{booktabs}
\usepackage{amsmath,amsfonts,amssymb}
\usepackage{subcaption}
\usepackage{multirow}
\usepackage{colortbl}
\usepackage{listings}
\usepackage{xparse}
\usepackage{fontawesome5}
\usepackage{threeparttable}

\graphicspath{{./}{fig/}{figures/}{plot/}{pdf/}{table/}}
\usepackage{amsthm}
\usepackage{tcolorbox}
\tcbuselibrary{skins,breakable}
\tcbuselibrary{listingsutf8}
\usepackage{titletoc}
\usepackage{pifont}
\usepackage{mathtools}
\usepackage{bbm}
\usepackage{makecell}
\usepackage{adjustbox}
\usepackage{xspace}

\newcommand{\best}[1]{\textbf{#1}}
\newcommand{\second}[1]{\underline{#1}}

\title{Optical Reasoning: Rethinking Images as an Expressive Reasoning Medium Beyond Text}
\runningtitle{Optical Reasoning}
\PublicDate{2026-06-08}

\author{%
  {\Authfont
    \textbf{Yutong Bian}\textsuperscript{1}\equal \quad
    \textbf{Dongjie Cheng}\textsuperscript{1}\equal \quad
    \textbf{Heming Xia}\textsuperscript{1} \quad
    \textbf{Yongqi Li}\textsuperscript{1}\advisor \quad
    \textbf{Wenjie Li}\textsuperscript{1}
  }\\
  {\Affilfont
    \textsuperscript{1} The Hong Kong Polytechnic University \quad \\
    \texttt{yutongbian02@gmail.com, dong-jie.cheng@connect.polyu.hk, liyongqi0@gmail.com}
  }
}

\begin{document}

\begin{abstract}
Chain-of-Thought (CoT) improves the performance of Large Language Models (LLMs) and has been extended to Multimodal Large Language Models (MLLMs). More recent work further moves from text-based multimodal reasoning toward interleaved-modal reasoning, where intermediate steps can incorporate both textual rationales and visual evidence. In this work, we propose a bolder and more ambitious idea: could images alone serve as the reasoning medium for both language and multimodal tasks? To explore this, we propose optical reasoning, which treats images as a standalone reasoning medium. We instantiate this concept with two variants: typographic-based optical reasoning, which optimizes visual layouts for compact rationale rendering, and graphical-based optical reasoning, which composes text and graphical elements into structured visual rationales. Across mathematical, scientific, and interleaved-modal reasoning benchmarks, optical reasoning can match or even exceed traditional text reasoning while reducing reasoning tokens by an average of 28.57\% on language tasks and 16\% on multimodal tasks, achieving \(1.96\times\) the token efficiency of text reasoning. These results show that images can effectively and efficiently encode rationales while providing a unified visual canvas for reasoning. The code is attached for reproducibility and subsequent open release.
\end{abstract}

\newcommand{\TitleLinks}{%
\centering
    \vspace{6pt}
    {\noindent\absfont\fontsize{11}{13}\selectfont
    \faGithub\ Code: \url{https://github.com/ModalityDance/Optical-Reasoning}\par}%
}

\maketitle


\input{1_intro}
\input{2_relatedwork}
\input{3_method}
\input{4_experiments}

\input{5_conclusion}


\bibliographystyle{unsrtnat} 
\bibliography{ref}


\appendix
\input{6_appendix}

\end{document}

%% file: 1_intro.tex
\section{Introduction}
\begin{figure}[t!]
\centering
  \includegraphics[width=1.0\linewidth]{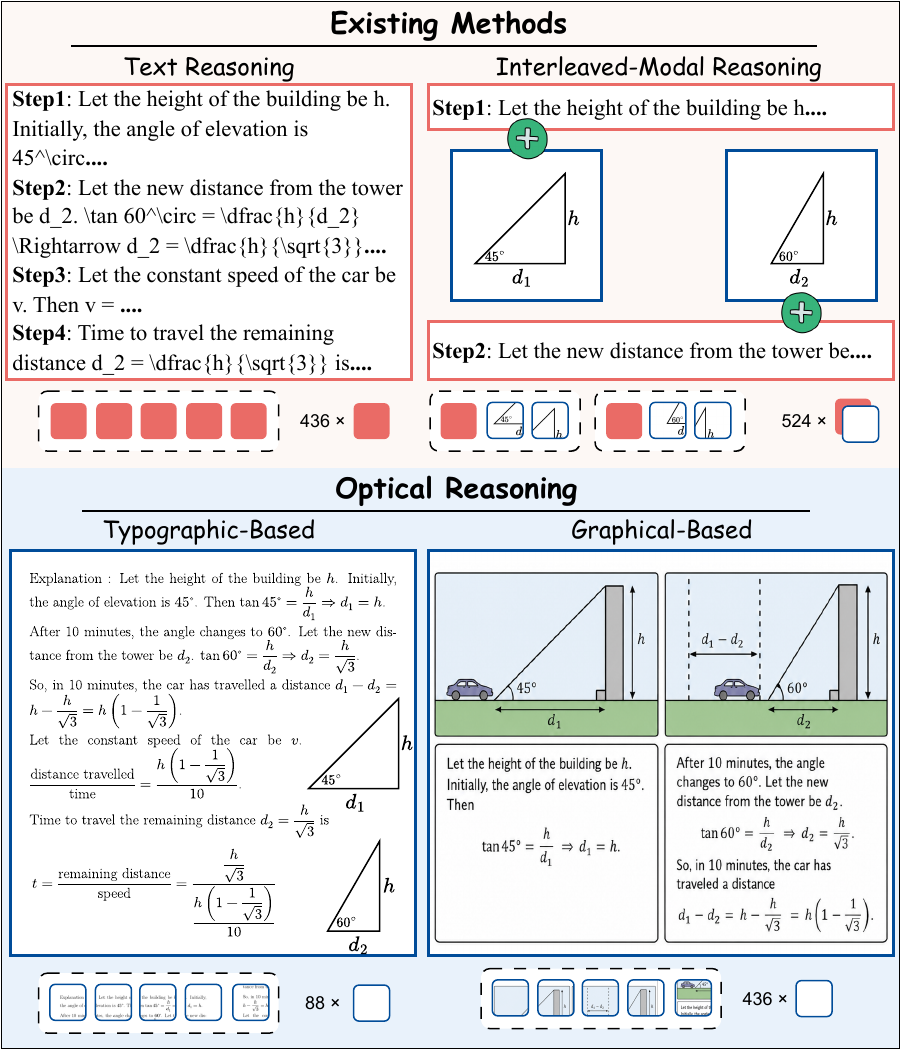}
  \caption{An example of different reasoning paradigms. Optical reasoning represents rationales as either dense typographic layouts or step-aligned graphical compositions. Dark blue boxes denote images, and light red boxes denote text.}
  \vspace{-1.5em}
  \label{fig:intro}
\end{figure}

Chain-of-Thought (CoT)~\citep{wei2023chainofthoughtpromptingelicitsreasoning} improves the performance of Large Language Models (LLMs) by eliciting intermediate reasoning steps before deriving final answers, as illustrated in Figure~\ref{fig:intro}. Recently, Large Reasoning Models (LRMs)~\citep{ deepseek-r1} further strengthen this paradigm by producing longer reasoning traces, leading to improved performance on complex reasoning tasks.

Building on the success of reasoning in LLMs, recent studies~\citep{mondal2024kamcotknowledgeaugmentedmultimodal, zhou2025r1zerosahamomentvisual, meng2025mmeurekaexploringfrontiersmultimodal} have extended CoT to multimodal reasoning, where Multimodal Large Language Models (MLLMs) answer questions by jointly interpreting textual prompts and visual inputs. In this setting, CoT improves performance by generating textual reasoning steps based on visual inputs and questions. More recent work further moves beyond purely textual rationales toward interleaved-modal reasoning~\citep{gao_interleaved-modal_2025, zheng_deepeyes_2026, cheng2026omnir1unifiedgenerativeparadigm}, where reasoning steps may incorporate both textual rationales and visual evidence. As illustrated in the interleaved-modal reasoning part of Figure~\ref{fig:intro}, this trend suggests that visual information could also contribute to the reasoning process.

Based on the above process, we arrived at a bolder and more ambitious idea: \textbf{could images alone serve as the reasoning medium for both language and multimodal tasks?} We believe this transition yields two primary potentials. 1) First, inspired by optical context compression~\citep{wei_deepseek-ocr_2025, shi_codeocr_2026, shi_memocr_2026, feng_agentocr_2026}, which transforms contexts into images to reduce input tokens, we believe that rendering rationales into images could maintain reasoning performance while substantially improving reasoning efficiency. 2) Second, given the critical role of visual information in multimodal tasks, images provide a unified visual canvas that can integrate text, graphical elements, and spatial layouts naturally.

In this work, we propose optical reasoning, which treats images as the sole reasoning medium, and instantiate this concept through two variants. 1) Typographic-based optical reasoning (T-OR), illustrated in the typographic part of Figure~\ref{fig:intro}, employs an optimal layout token strategy designed to maximize information density under a controllable set of reasoning tokens. Specifically, it searches over text width and font size to improve canvas utilization while preserving readability and complete reasoning content. 2) Graphical-based optical reasoning (G-OR), illustrated in the graphical part of Figure~\ref{fig:intro}, employs a step-aligned composition strategy to unify textual rationales and graphical elements within an image. Specifically, G-OR decomposes each rationale into reasoning steps and assigns each step to a corresponding visual panel, where concepts and relations are represented as graphical structures while key formulas and text are retained as explicit reasoning anchors. Together, these two variants examine images as both compact carriers and unified visual canvases for reasoning.

We conducted experiments across mathematical reasoning, scientific reasoning, and interleaved-modal reasoning benchmarks using five frontier MLLMs: GPT-5.1~\citep{singh2026openaigpt5card}, Gemini 2.5 Flash~\citep{comanici2025gemini25pushingfrontier}, Claude Sonnet 4.5\footnote{\url{https://www-cdn.anthropic.com/963373e433e489a87a10c823c52a0a013e9172dd.pdf}}, Kimi K2.5~\citep{kimiteam2026kimik25visualagentic}, and Qwen3-VL-235B~\citep{bai2025qwen3vltechnicalreport}. For language tasks, T-OR matches or exceeds text reasoning in seven model-benchmark pairs while reducing reasoning tokens by an average of 28.57\%. In the remaining cases where T-OR underperforms text reasoning, its best setting trails by an average accuracy gap of 0.027 while still reducing the number of reasoning tokens by 20\%. For multimodal tasks, T-OR matches or exceeds text reasoning in five model-benchmark pairs while reducing reasoning tokens by an average of 16\%; when T-OR falls behind, the average accuracy gap is only 0.014 with a 32\% token reduction. More broadly, under the Marginal Accuracy Gain (MAG) metric in Eq.~\ref{eq:mag}, each visual reasoning token achieves \(1.96\times\) the efficiency of a text reasoning token. These results validate that images enable effective and efficient compression of interleaved-modal rationales.

The contributions are summarized as follows
\begin{itemize}
    \item We introduce optical reasoning, positioning images as a promising reasoning medium that not only compactly encodes textual rationales but also offers a unified visual canvas to seamlessly integrate text and graphical elements.
    \item We instantiate optical reasoning with typographic-based and graphical-based variants to explore the efficacy of images as a reasoning medium. Specifically, the former employs an optimized layout token strategy to maximize information density, while the latter leverages the complementarity between symbolic derivations and spatial relations to enable unified multimodal reasoning.
    \item We evaluate optical reasoning across 5 benchmarks and 5 advanced MLLMs, demonstrating that images are an effective and efficient medium with unique capabilities for structuring rationales, achieving \(1.96\times\) the token efficiency of text reasoning.
\end{itemize}

%% file: 2_relatedwork.tex
\section{Related Work}
\paragraph{Interleaved-modal reasoning.}
Multimodal reasoning is now shifting from text reasoning toward interleaved-modal reasoning. ICoT~\citep{gao_interleaved-modal_2025} introduced the concept of interleaved-modal reasoning, where the model identifies relevant visual regions via attention and integrates them into the reasoning process. MINT-CoT~\citep{chen2025mintcotenablinginterleavedvisual} further introduced interleaved visual tokens into mathematical CoT, enabling fine-grained visual grounding during reasoning. DeepEyes~\citep{zheng_deepeyes_2026} employed active perception with external tools. MVoT~\citep{li2025imaginereasoningspacemultimodal} enabled visual thinking by generating image visualizations of rationales. Zebra-CoT~\citep{li2025zebracotdatasetinterleavedvision} further provided a large-scale interleaved-modal dataset. Omni-R1~\citep{cheng2026omnir1unifiedgenerativeparadigm} proposed a unified generative paradigm for multimodal reasoning by generating intermediate images during reasoning. These methods enrich text-based reasoning with visual information, yet still rely primarily on textual rationales; in contrast, we explore images as standalone reasoning media.
\paragraph{Optical compression.}
Images have recently shown strong potential for compressing textual content. DeepSeek-OCR~\citep{wei_deepseek-ocr_2025} formalized this idea as context optical compression, where long textual contexts are encoded as compact images. Glyph~\citep{cheng2025glyphscalingcontextwindows} further explored optical compression for scaling context windows. Following this paradigm, CodeOCR~\citep{shi_codeocr_2026}, AgentOCR~\citep{feng_agentocr_2026}, and MemOCR~\citep{shi_memocr_2026} extend optical compression to more scenarios. Moreover, VTC-R1~\citep{wang2026vtcr1visiontextcompressionefficient} incorporated optical compression into reasoning by rendering previous textual rationales into images and feeding them back to MLLMs. RoT~\citep{wang2026renderofthoughtrenderingtextualchainofthought} similarly rendered textual rationales into images, for training the latent reasoning. These studies mainly treat images as a compression of textual rationales, whereas we explore images as standalone reasoning media that can organize text, graphical elements, and spatial layouts.

%% file: 3_method.tex
\section{Method}
In this section, we formalize text reasoning, introduce optical reasoning, and instantiate it using typographic-based and graphical-based variants.
\subsection{Preliminary: Text Reasoning}
In standard text reasoning~\citep{wei2023chainofthoughtpromptingelicitsreasoning}, language models generate intermediate rationales and final answers in text. We assume the rationale is available, either generated by a model or provided externally, and focus on how different media represent the same reasoning content.
Let \(\mathbf{q}=(q_1,\dots,q_{N_q})\) denote its tokenized question. Let \(\mathbf{r}_{\mathrm{txt}}\) denote the textual rationale sequence:
\begin{equation}
\mathbf{r}_{\mathrm{txt}}
=
(r^{\mathrm{txt}}_1,\dots,r^{\mathrm{txt}}_{N_{\mathrm{txt}}}),
\end{equation}
where each \(r^{\mathrm{txt}}_i\) is a textual rationale unit, such as a text span or an equation, and \(N_{\mathrm{txt}}\) is the number of textual rationale units. For model inference, \(\mathbf{r}_{\mathrm{txt}}\) is serialized into text tokens. Given \(\mathbf{q}\) and \(\mathbf{r}_{\mathrm{txt}}\), the model \(\pi_{\theta}\) decodes the answer token sequence \(\mathbf{a}\):
\begin{equation}
\mathbf{a} \sim \pi_{\theta}
\left(\cdot \mid \mathbf{q}, \mathbf{r}_{\mathrm{txt}}\right).
\end{equation}
Thus, \(\mathbf{r}_{\mathrm{txt}}\) serves as the reasoning medium.
\subsection{Optical Reasoning}
\label{subsec:OR}
Motivated by the potential of images to compactly replace rationales and to unify textual, graphical, and spatial information, optical reasoning represents intermediate rationales as images and uses them as the reasoning medium. To support both textual and visual rationales, we define a unified interleaved-modal rationale sequence as
\begin{equation}
\mathbf{r}_{\mathrm{mix}}
=
(r^{\mathrm{mix}}_1,\dots,r^{\mathrm{mix}}_{N_{\mathrm{mix}}}),
\quad
r^{\mathrm{mix}}_i\in\{r^{\mathrm{txt}},r^{\mathrm{vis}}\},
\end{equation}
where $r^{\mathrm{mix}}_i$ denotes the $i$-th rationale unit and $N_{\mathrm{mix}}$ represents the total number of interleaved rationale units. A rationale unit is textual when $r^{\mathrm{mix}}_i = r^{\mathrm{txt}}$, encompassing text spans or equations, and visual when $r^{\mathrm{mix}}_i = r^{\mathrm{vis}}$, which includes images, diagrams, or visual evidence segments. When all rationale units equal $r^{\mathrm{txt}}$, the sequence $\mathbf{r}_{\mathrm{mix}}$ corresponds to the textual rationale sequence $\mathbf{r}_{\mathrm{text}}$ utilized in text reasoning.
A renderer \(g\) maps the interleaved-modal rationale into an image:
\begin{equation}
I = g(\mathbf{r}_{\mathrm{mix}}).
\end{equation}
Let \(N_{\mathrm{vis}}\) be the number of visual reasoning tokens. The visual encoder \(\phi(\cdot)\) maps \(I\) to
\begin{equation}
\mathbf{z}_{\mathrm{vis}}
=
\phi(I)
=
(z^{\mathrm{vis}}_1,z^{\mathrm{vis}}_2,\dots,z^{\mathrm{vis}}_{N_{\mathrm{vis}}}),
\end{equation}
where \(z^{\mathrm{vis}}_i\) is the \(i\)-th visual reasoning token. Under optical reasoning, the model derives the answer from question tokens and visual reasoning tokens:
\begin{equation}
\mathbf{a} \sim \pi_{\theta}
\left(\cdot \mid \mathbf{q}, \mathbf{z}_{\mathrm{vis}}\right).
\end{equation}
Thus, optical reasoning represents rationales with image tokens rather than text tokens. Based on this formulation, we instantiate optical reasoning with two renderers: a typographic renderer \(g_{\mathrm{typo}}\) and a graphical renderer \(g_{\mathrm{graph}}\).
\paragraph{Typographic-based optical reasoning.}
For T-OR, we render the interleaved-modal rationale sequence \(\mathbf{r}_{\mathrm{mix}}\) into a compact typographic image under a controllable reasoning-token budget \(B\). The renderer \(g_{\mathrm{typo}}\), implemented with XeLaTeX\footnote{\url{https://xetex.sourceforge.net/}}, preserves the original order of rationale units: textual units are typeset as text, equations, or tables, while visual units are inserted as image blocks. Formally, it maps the rationale sequence into a typographic rationale image:
\begin{equation}
I_{\mathrm{typo}}
=
g_{\mathrm{typo}}(\mathbf{r}_{\mathrm{mix}};\ell^\star).
\end{equation}
Here, \(\ell^\star\) is the selected layout configuration. We define a layout configuration as
\begin{equation}
\ell=(w,s,\gamma,p),
\end{equation}
where \(w\) denotes the text width, \(s\) denotes the font size, \(\gamma\) denotes the line spacing, and \(p\) denotes the page padding. These variables control how rationale units are placed on the image canvas.
In our implementation, \(g_{\mathrm{typo}}\) searches over the candidate width set \(\mathcal{W}\) and font-size set \(\mathcal{S}\), while using default values \(\gamma_0\) and \(p_0\) for line spacing and page padding. Therefore, the candidate layout set is
\begin{equation}
\mathcal{C}
=
\{(w,s,\gamma_0,p_0)\mid w\in\mathcal{W},\, s\in\mathcal{S}\}.
\end{equation}
Given the budget \(B\), the renderer searches for the most compact and readable layout \(\ell^\star\). The feasible layout set under this budget is defined as
\begin{equation}
\mathcal{C}_B
=
\left\{
\ell\in\mathcal{C}
\mid
N_\mathrm{vis}\le B
\right\}.
\end{equation}
Each feasible candidate image \(I_{\mathrm{typo}}\) is evaluated by a layout score:
\begin{equation}
S(\ell)
=
\rho(\ell)-\lambda\epsilon(\ell),
\end{equation}
where \(\rho(\ell)\) is the fill ratio, \(\epsilon(\ell)\) is the layout penalty, and \(\lambda\) is the penalty weight. Specifically, \(\rho(\ell)\) measures the ratio between the occupied content and the canvas, while \(\epsilon(\ell)\) aggregates penalties for excessive margins and overly tight layouts that reduce readability. The optimal layout is selected as
\begin{equation}
\ell^\star
=
\arg\max_{\ell\in\mathcal{C}_B}
S(\ell).
\end{equation}
In practice, the renderer first performs a coarse search over \(\mathcal{C}\) with a font-size step \(\delta_s\), while retaining only candidates that satisfy the reasoning-token budget. If a feasible candidate layout reaches the threshold \(\tau_{\min}\), the renderer further examines neighboring font sizes skipped during the coarse search. If no feasible candidate reaches \(\tau_{\min}\), the renderer falls back to the feasible candidate with the highest score. This strategy preserves the full rationale content while reducing redundant visual space under the target reasoning-token budget.
\paragraph{Graphical-based optical reasoning.}
Graphical-based optical reasoning transforms \(\mathbf{r}_{\mathrm{mix}}\) into a unified image-based rationale that organizes reasoning with text, graphical elements, and spatial layouts. We define the graphical renderer \(g_{\mathrm{graph}}\) as
\begin{equation}
I_{\mathrm{graph}}
=
g_{\mathrm{graph}}(\mathbf{r}_{\mathrm{mix}}),
\end{equation}
where \(\mathbf{r}_{\mathrm{mix}}\) is the interleaved-modal rationale sequence.
We instantiate \(g_{\mathrm{graph}}\) with Nano Banana 2\footnote{\url{https://storage.googleapis.com/deepmind-media/Model-Cards/Gemini-3-1-Flash-Image-Model-Card.pdf}} via a structured prompt. The prompt provides the problem, the rationale, and optional visual images, and asks the renderer to produce a multi-panel graphical rationale following a step-aligned composition strategy. Specifically, the renderer decomposes the rationale into reasoning steps and assigns each step to a corresponding visual panel. Key reasoning text, equations are preserved as textual annotations, while graphical elements and spatial layouts are used to reorganize the rationale visually. The full prompt template is provided in Appendix~\ref{appendix:t2i}.
The resulting graphical rationale image \(I_{\mathrm{graph}}\) is encoded by \(\phi(\cdot)\) into visual reasoning tokens \(\mathbf{z}_{\mathrm{vis}}\) for answer derivation. Compared with \(I_{\mathrm{typo}}\), which preserves the original order of rationale units, \(I_{\mathrm{graph}}\) reorganizes the rationale within a unified visual canvas.
\input{tables/typographic_exp}
\begin{figure*}[t!]
    \centering
    \includegraphics[width=\textwidth]{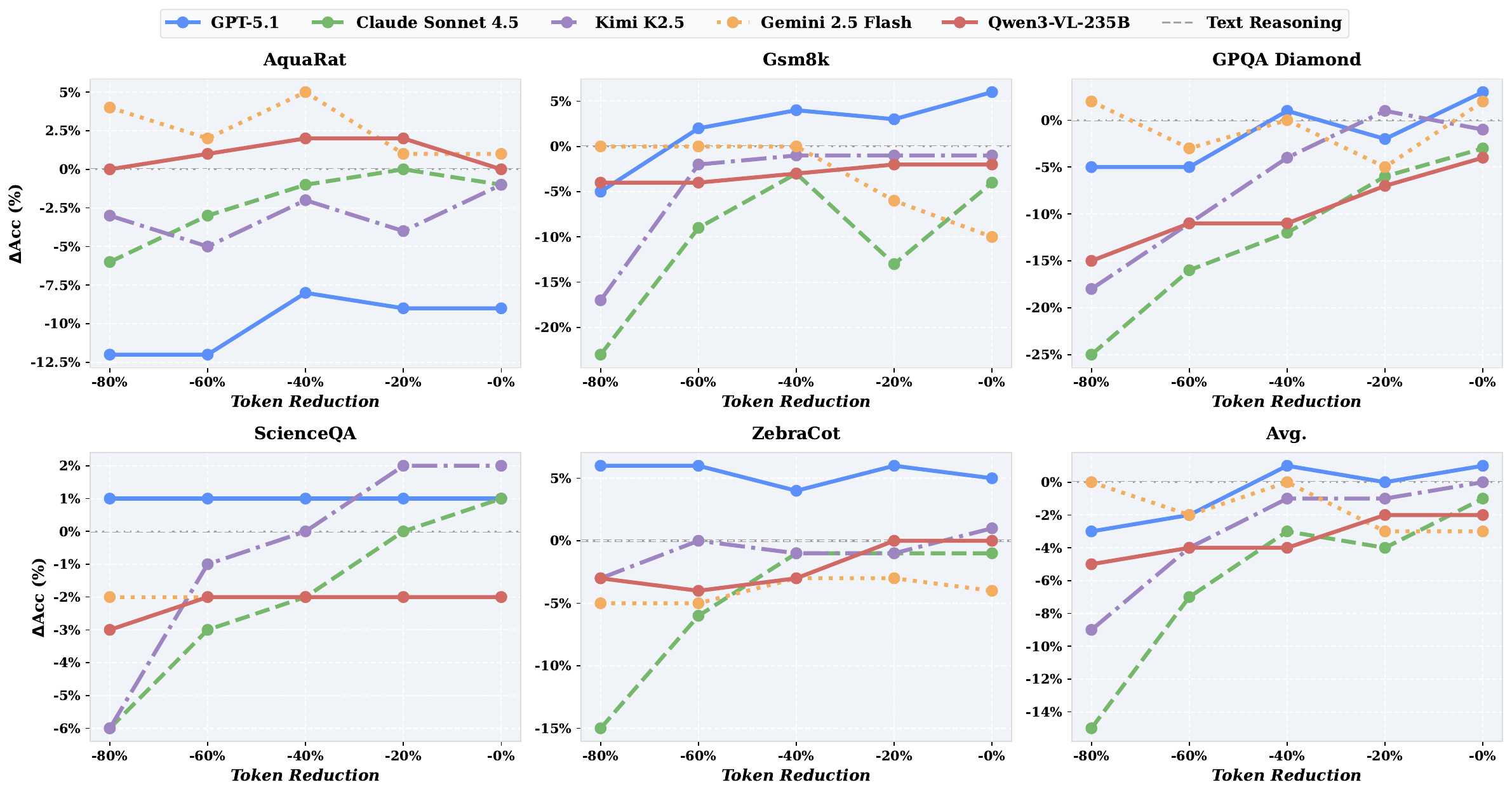}
    \caption{Analysis on token compression ratios. We report the accuracy change of T-OR over text reasoning under different reasoning tokens. Positive values indicate improvements over the text reasoning baseline.}
    \vspace{-1em}
    \label{fig:main-exp-token-reduction}
\end{figure*}

%% file: tables/typographic_exp.tex
\begin{table*}[!t]
\small
\centering
\renewcommand{\arraystretch}{0.88}
\resizebox{\textwidth}{!}{
\begin{tabular}{c|l|c|c|cc|cc|c|c|c}
\toprule
& \multirow{3}{*}{\textbf{Setting}} & \multirow{3}{*}{\textbf{\begin{tabular}[c]{@{}c@{}}Reasoning \\ Tokens\end{tabular}}} & \multirow{3}{*}{\textbf{\begin{tabular}[c]{@{}c@{}}Token \\ Reduction\end{tabular}}} & \multicolumn{2}{c|}{\textbf{Mathematical}} & \multicolumn{2}{c|}{\textbf{Scientific}} & \textbf{Multimodal} & \multirow{3}{*}{\textbf{Avg. Acc.}} & \multirow{3}{*}{\textbf{MAG}} \\
\cmidrule(lr){5-6} \cmidrule(lr){7-8} \cmidrule(lr){9-9}
& & & & \textbf{AquaRat} & \textbf{Gsm8k} & \textbf{\begin{tabular}[c]{@{}c@{}}GPQA \\ Diamond\end{tabular}} & \textbf{ScienceQA} & \textbf{Zebra-Cot} & & \\
\midrule

\multirow{7}{*}{\rotatebox{90}{GPT-5.1}} 
& No reasoning & 0.0 & -100\% & 0.4961 & 0.3222 & 0.4646 & 0.6737 & 0.2633 & 0.4440 & -- \\
\cmidrule(lr){2-11}
& Text reasoning & 257.8 & -0\% & \best{0.6811} & 0.8908 & 0.7475 & 0.9646 & 0.5733 & 0.7715 & 1.27 \\
\cmidrule(lr){2-11}
& \multirow{5}{*}{\textbf{T-OR}} & 51.8 & -80\% & 0.5610 & 0.8446 & 0.6970 & \second{0.9695} & \best{0.6300} & 0.7404 & \best{5.72} \\
& & 103.7 & -60\% & 0.5630 & 0.9098 & 0.7020 & 0.9673 & \second{0.6267} & 0.7538 & \second{2.99} \\
& & 155.6 & -40\% & \second{0.6044} & \second{0.9334} & \second{0.7626} & 0.9690 & 0.6133 & \second{0.7765} & 2.14 \\
& & 207.4 & -20\% & 0.5866 & 0.9249 & 0.7273 & 0.9679 & \second{0.6267} & 0.7667 & 1.56 \\
& & 259.3 & -0\% & 0.5866 & \best{0.9500} & \best{0.7778} & \best{0.9739} & 0.6167 & \best{0.7810} & 1.30 \\
\midrule

\multirow{7}{*}{\rotatebox{90}{Kimi K2.5}} 
& No reasoning & 0.0 & -100\% & 0.6969 & 0.6179 & 0.4646 & 0.6411 & 0.3500 & 0.5541 & -- \\
\cmidrule(lr){2-11}
& Text reasoning & 257.8 & -0\% & \best{0.7677} & \best{0.9962} & \second{0.7525} & 0.9314 & 0.7000 & \second{0.8296} & 1.07 \\
\cmidrule(lr){2-11}
& \multirow{5}{*}{\textbf{T-OR}} & 51.8 & -80\% & 0.7441 & 0.8340 & 0.5657 & 0.8720 & 0.6733 & 0.7378 & \best{3.55} \\
& & 103.7 & -60\% & 0.7165 & 0.9757 & 0.6364 & 0.9194 & \second{0.7033} & 0.7903 & \second{2.28} \\
& & 155.6 & -40\% & 0.7520 & 0.9894 & 0.7071 & 0.9341 & 0.6933 & 0.8152 & 1.68 \\
& & 207.4 & -20\% & 0.7283 & 0.9924 & \best{0.7626} & \second{0.9455} & 0.6933 & 0.8245 & 1.30 \\
& & 259.3 & -0\% & \second{0.7598} & \second{0.9932} & 0.7424 & \best{0.9504} & \best{0.7133} & \best{0.8318} & 1.07 \\
\midrule

\multirow{7}{*}{\rotatebox{90}{Gemini 2.5 Flash}} 
& No reasoning & 0.0 & -100\% & 0.6890 & 0.5967 & 0.4293 & 0.6737 & 0.2600 & 0.5195 & -- \\
\cmidrule(lr){2-11}
& Text reasoning & 257.8 & -0\% & 0.7323 & \best{0.9947} & \second{0.7626} & \best{0.9782} & \best{0.6867} & \best{0.8309} & 1.21 \\
\cmidrule(lr){2-11}
& \multirow{5}{*}{\textbf{T-OR}} & 51.8 & -80\% & \second{0.7677} & 0.9894 & \best{0.7828} & 0.9597 & 0.6400 & \second{0.8279} & \best{5.95} \\
& & 103.7 & -60\% & 0.7520 & \second{0.9909} & 0.7323 & 0.9592 & 0.6367 & 0.8142 & \second{2.84} \\
& & 155.6 & -40\% & \best{0.7835} & 0.9894 & \second{0.7626} & 0.9592 & \second{0.6600} & \best{0.8309} & 2.00 \\
& & 207.4 & -20\% & 0.7402 & 0.9310 & 0.7071 & \second{0.9613} & \second{0.6600} & 0.7999 & 1.35 \\
& & 259.3 & -0\% & 0.7362 & 0.8923 & \best{0.7828} & \second{0.9613} & 0.6467 & 0.8039 & 1.10 \\
\midrule

\multirow{7}{*}{\rotatebox{90}{Qwen3-VL-235B}} 
& No reasoning & 0.0 & -100\% & 0.6024 & 0.4193 & 0.4141 & 0.7102 & 0.2800 & 0.4852 & -- \\
\cmidrule(lr){2-11}
& Text reasoning & 257.8 & -0\% & 0.7441 & \best{0.9970} & \best{0.7727} & \best{0.9668} & \best{0.6600} & \best{0.8281} & 1.33 \\
\cmidrule(lr){2-11}
& \multirow{5}{*}{\textbf{T-OR}} & 51.8 & -80\% & 0.7362 & 0.9568 & 0.6212 & 0.9423 & 0.6267 & 0.7766 & \best{5.63} \\
& & 103.7 & -60\% & 0.7520 & 0.9553 & 0.6566 & \second{0.9526} & 0.6200 & 0.7873 & \second{2.91} \\
& & 155.6 & -40\% & \best{0.7638} & 0.9704 & 0.6616 & 0.9477 & 0.6300 & 0.7947 & 1.99 \\
& & 207.4 & -20\% & \second{0.7559} & \second{0.9803} & 0.7020 & 0.9477 & \second{0.6567} & 0.8085 & 1.56 \\
& & 259.3 & -0\% & 0.7441 & 0.9795 & \second{0.7323} & 0.9499 & \second{0.6567} & \second{0.8125} & 1.26 \\
\midrule

\multirow{7}{*}{\rotatebox{90}{Claude 4.5}} 
& No reasoning & 0.0 & -100\% & 0.7756 & 0.7384 & 0.5303 & 0.5294 & 0.3300 & 0.5807 & -- \\
\cmidrule(lr){2-11}
& Text reasoning & 257.8 & -0\% & \best{0.8465} & \best{0.9947} & \best{0.8434} & \second{0.9178} & \best{0.6700} & \best{0.8545} & 1.06 \\
\cmidrule(lr){2-11}
& \multirow{5}{*}{\textbf{T-OR}} & 51.8 & -80\% & 0.7913 & 0.7589 & 0.5909 & 0.8557 & 0.5200 & 0.7034 & \best{2.37} \\
& & 103.7 & -60\% & 0.8189 & 0.8999 & 0.6768 & 0.8922 & 0.6100 & 0.7795 & \second{1.92} \\
& & 155.6 & -40\% & \second{0.8386} & \second{0.9591} & 0.7222 & 0.9031 & 0.6600 & 0.8166 & 1.52 \\
& & 207.4 & -20\% & \best{0.8465} & 0.8597 & \second{0.7828} & 0.9161 & \second{0.6633} & 0.8137 & 1.12 \\
& & 259.3 & -0\% & \second{0.8386} & 0.9515 & \second{0.8131} & \best{0.9254} & 0.6600 & \second{0.8377} & 0.99 \\
\bottomrule
\end{tabular}
}
\caption{Typographic-based optical reasoning(T-OR) evaluation results across five benchmarks. The Reasoning Tokens reports the average usage per sample across all five benchmarks. MAG reports the marginal accuracy gain over the no-reasoning baseline per 1,000 reasoning tokens. Bold and underlined values indicate the best and second-best results within each comparison group, respectively.}
\label{tab:main-exp}
\end{table*}

%% file: 4_experiments.tex
\section{Experiments}
\subsection{Experimental Setup}
\paragraph{Datasets and evaluation.}
We evaluate optical reasoning on three reasoning categories: mathematical reasoning with AquaRat~\citep{DBLP:journals/corr/LingYDB17} and GSM8K~\citep{cobbe2021gsm8k}, scientific reasoning with GPQA Diamond~\citep{rein2023gpqagraduatelevelgoogleproofqa} and ScienceQA~\citep{lu2022learnexplainmultimodalreasoning}, and interleaved-modal reasoning with Zebra-CoT~\citep{li2025zebracotdatasetinterleavedvision}. Detailed descriptions of these datasets are provided in the Appendix~\ref{appendix:bench}.
We used five frontier MLLMs for our evaluation. Specifically, we selected three closed-source models (GPT-5.1~\citep{singh2026openaigpt5card}, Gemini 2.5 Flash~\citep{comanici2025gemini25pushingfrontier}, and Claude Sonnet 4.5) and two open-source models (Kimi K2.5~\citep{kimiteam2026kimik25visualagentic}, Qwen3-VL-235B~\citep{bai2025qwen3vltechnicalreport}).
We use accuracy as the primary metric, first applying rule-based matching and then using a large language model judge for unmatched cases with the prompt in Appendix~\ref{appendix:llm_judge}.To better measure token efficiency, we utilized the Marginal Accuracy Gain (MAG) per reasoning token. Specifically, MAG is defined as the accuracy improvement over the no reasoning baseline, normalized by the number of reasoning tokens, capturing the accuracy gain per reasoning token.
\begin{equation}
\label{eq:mag}
\mathrm{MAG}_{m} = \frac{Acc_{m} - Acc_{\mathrm{no}}}{N_m}, 
\quad m \in \{\mathrm{text}, \mathrm{visual}\}.
\end{equation}
\paragraph{Baselines.}
We compared two reasoning settings. \textit{No reasoning} provides only the task input and measures the model's direct-answer capability. \textit{Text reasoning} provides the textual rationale and serves as the upper bound in our setup. 
\paragraph{Implementation.}
In this work, we primarily used rationales from open-source CoT datasets to assess the effectiveness of optical reasoning. Nevertheless, externally provided rationales may also affect model behavior, thereby confounding the assessment of end-to-end reasoning performance. To examine whether optical reasoning remains effective when models generate their own rationales, we further evaluated optical reasoning in Section~\ref{para:self-generated-rationale}.
To ensure a fair and controlled evaluation, we strictly controlled the reasoning image resolution across all models according to the patch-to-token counting rules described in Appendix~\ref{appendix:patch-token-mapping}. We also prevented answer leakage by removing explicit answers from the rationales, thereby requiring models to infer the solutions independently. The evaluation prompts for all reasoning settings are provided in Appendix~\ref{appendix:evaluation-prompts}. In addition, we disabled the reasoning mode so that models rely solely on given rationales. For inference, we set the temperature to 0 for all models except Kimi K2.5~\citep{kimiteam2026kimik25visualagentic}, which only supports a fixed temperature of 0.6 in the non-thinking mode.
\subsection{Results on Typographic-based Optical Reasoning}
Through the typographic-based variant, we examined whether images can serve as a more compact but effective medium for rationales. 
As shown in the Table~\ref{tab:main-exp}, we observe two main trends across diverse benchmarks and models. First, under comparable reasoning tokens, typographic-based optical reasoning matches or outperforms text reasoning for multiple model-benchmark pairs, while achieving an average MAG \(1.96\times\) higher than that of text reasoning across all token-budget ratios. Second, on Zebra-CoT~\citep{li2025zebracotdatasetinterleavedvision}, a benchmark requiring interleaved-modal reasoning, T-OR outperforms text reasoning on Gemini 2.5 Flash~\citep{comanici2025gemini25pushingfrontier}, Kimi K2.5~\citep{kimiteam2026kimik25visualagentic}, and GPT-5.1~\citep{singh2026openaigpt5card}, and remains within 1\% of performance drop relative to text reasoning on the remaining two models. By integrating rationales and multimodal evidence into images, T-OR demonstrates that images can serve as an effective and efficient reasoning medium. Figure~\ref{fig:main-exp-token-reduction} further shows how models respond to reasoning token compression. The performance of Gemini 2.5 Flash~\citep{comanici2025gemini25pushingfrontier} remains competitive even under aggressive compression. In contrast, Kimi K2.5~\citep{kimiteam2026kimik25visualagentic} and Claude Sonnet 4.5 improve more consistently as visual tokens increase. This phenomenon indicates that sensitivity to visual information density differs across model families. Moreover, the loss of visual details caused by optical 2D mapping does not necessarily imply that the cues required for reasoning are lost, at least for certain models. We further analyzed this view in Section \ref{sec:extreme}.
\subsection{Results on Graphical-based Optical Reasoning}
Building on prior experiments demonstrating that images can serve as a better reasoning medium than text, we further explored the unique advantages of the image with our graphical-based optical reasoning on AquaRat~\citep{DBLP:journals/corr/LingYDB17}.
It is shown in Table~\ref{tab:graphical-optical-reasoning} that graphical-based optical reasoning achieves the best accuracy, outperforming both text reasoning and typographic-based optical reasoning. By integrating text, graphical elements, and spatial layouts within a unified visual canvas, graphical-based optical reasoning can convey more complex multimodal evidence naturally. This suggests that images are not merely compact containers for text but an expressive reasoning medium. We further illustrate this advantage in Section~\ref{subsec:case_study}.
\input{tables/graphical_exp}
\subsection{Ablation Study}
Since typographic-based optical reasoning relies on a text-to-visual rendering engine, we investigated the impact of key factors during the rendering process by evaluating the performance of the GPT-5.1~\citep{singh2026openaigpt5card} model on the GPQA Diamond~\citep{rein2023gpqagraduatelevelgoogleproofqa} dataset. Table~\ref{tab:ablation_exp} presents the detailed experimental results.
\paragraph{Effects of layout style.} We ablated color and font family by varying each rendering factor independently while keeping the others fixed. As shown in Table~\ref{tab:ablation_exp}, red achieves the highest accuracy, outperforming the black baseline, whereas green leads to the lowest accuracy. For font family, "Heros" yields the best performance. These results indicate that visual appearance affects how effectively the model decodes typographic rationales in optical reasoning. A possible reason is that high-contrast colors and clean fonts make key textual regions easier for the visual encoder to identify and parse.
\paragraph{Effects of layout density.}
We further studied font size and text width to examine the effect of layout density on visual reasoning. As shown in Table~\ref{tab:ablation_exp}, very small fonts substantially reduce accuracy, while moderate font sizes perform better; meanwhile, a narrower text width achieves higher accuracy than wider layouts. These observations suggest that optical reasoning benefits from compact but readable visual layouts. This may be because overly dense layouts impair legibility, while excessively wide layouts spread related reasoning steps too far apart, making it harder for the model to attend to them effectively.
\input{tables/ablation}
\subsection{In-depth Analysis}
\paragraph{Extreme compression.}
\label{sec:extreme}
To assess the robustness of optical reasoning under extreme token compression, we evaluated T-OR on Gemini 2.5 Flash~\citep{comanici2025gemini25pushingfrontier} with the AquaRat~\citep{DBLP:journals/corr/LingYDB17} dataset. We progressively reduced the number of visual tokens, pushing the average estimated token budget to as few as 1.2 tokens per example.
Table~\ref{tab:extreme-compression} demonstrates that optical reasoning remains effective even under extreme compression conditions. At a token budget ratio of $-98.75\%$, the model still achieves an accuracy that can outperform the baseline without reasoning. Optimal performance occurs at an average budget of only 7.2 reasoning tokens per example, which surpasses the performance of both text reasoning and the full-budget optical reasoning. These results indicate that optical reasoning does not rely exclusively on high-resolution visual legibility. Instead, compact visual layouts can preserve coarse yet informative reasoning cues within highly limited token budgets.
\input{tables/extreme_compression}
\paragraph{Impact of renderers.} 
\input{tables/render}
To assess how renderer choices affect optical reasoning, we conducted a controlled study on GPQA Diamond~\citep{rein2023gpqagraduatelevelgoogleproofqa} with four MLLMs~\citep{kimiteam2026kimik25visualagentic, comanici2025gemini25pushingfrontier, bai2025qwen3vltechnicalreport}. Specifically, we evaluated T-OR under three representative rendering engines: Pillow\footnote{\url{https://github.com/python-pillow/Pillow}}, Matplotlib~\citep{barrett2005matplotlib}, and XeLaTeX. We fixed the reasoning tokens, rendering strategy, and reasoning content, and varied only the backend used to render rationales into images, thereby isolating the renderer's effect.
As shown in Table~\ref{tab:render_exp}, different models prefer different rendering engines. Qwen3-VL~\citep{bai2025qwen3vltechnicalreport} and Claude achieve their best results with XeLaTeX, respectively, whereas Gemini~\citep{comanici2025gemini25pushingfrontier} performs best with Matplotlib. This indicates that the compatibility between rendering engines and MLLMs also affects the performance of optical reasoning, which may stem from MLLMs' varying abilities to interpret different visual styles.
\paragraph{Comparison with efficient text reasoning.} 
To examine whether optical reasoning offers advantages over existing efficient text reasoning methods, we compared it with LLMLingua-2~\citep{pan2024llmlingua2datadistillationefficient}, a representative text truncation method for reasoning compression, under equivalent reasoning tokens and the same rationales. We conducted this study on AquaRat~\citep{DBLP:journals/corr/LingYDB17} using Gemini-2.5 Flash~\citep{comanici2025gemini25pushingfrontier}. 
As shown in Table~\ref{tab:efficient-text-reasoning}, optical reasoning consistently outperforms LLMLingua-2~\citep{pan2024llmlingua2datadistillationefficient}.
This suggests that, unlike text truncation, optical reasoning can effectively preserve rationale content, because detail loss from optical 2D mapping does not necessarily remove the cues certain models need for reasoning.
\input{tables/efficient_text_reasoning}
\paragraph{Analysis on model-generated rationales}
\label{para:self-generated-rationale}
To verify whether optical reasoning generalizes to model-generated rationales, we evaluated GPT-5.1~\citep{singh2026openaigpt5card} on the GPQA Diamond dataset~\citep{rein2023gpqagraduatelevelgoogleproofqa}. The free reasoning baseline refers to text reasoning in which rationales are generated by MLLMs themselves. 
As shown in the table~\ref{tab:free-exp}, optical reasoning achieves comparable or superior performance relative to the free reasoning baseline. This confirms that optical reasoning holds substantial practical value in realistic reasoning scenarios rather than limited settings.
\input{tables/free_reasoning}

%% file: tables/graphical_exp.tex
\begin{table}[!t]
\small
\centering
\small
\setlength{\tabcolsep}{4pt}
\begin{tabular}{l|c|c}
\toprule
\textbf{Setting} & \textbf{\begin{tabular}[c]{@{}c@{}}Token \\ Reduction\end{tabular}} & \textbf{Acc.} \\
\midrule
No reasoning  & -100\% & 0.6890 \\
\cmidrule(lr){1-3}
Text reasoning & -0\% & 0.7323 \\
\cmidrule(lr){1-3}
\multirow{5}{*}{\textbf{T-OR}} & -80\% & 0.7677 \\
& -60\% & 0.7520 \\
& -40\% & \second{0.7835} \\
& -20\% & 0.7402 \\
& -0\% & 0.7362 \\
\cmidrule(lr){1-3}
\textbf{G-OR}& - & \best{0.8150} \\
\bottomrule
\end{tabular}
\caption{Accuracy on AquaRat under no reasoning, text reasoning, T-OR, and G-OR. Bold and underlined values indicate the best and second-best results.}
\label{tab:graphical-optical-reasoning}
\vspace{-1em}
\end{table}

%% file: tables/ablation.tex
\begin{table}[!t]
\small
\centering
\begin{tabular}{llc}
\toprule
\textbf{Factor} & \textbf{Value} & \textbf{Acc.} \\
\midrule
\multirow{4}{*}{Color}
& Black & 0.7727 \\
& Blue & \underline{0.7778} \\
& Green & 0.7525 \\
& Red & \textbf{0.7929} \\
\midrule
\multirow{4}{*}{Font Family}
& Heros & \textbf{0.7778} \\
& Latin Modern & 0.7626 \\
& Pagella & \underline{0.7626} \\
& Termes & 0.7576 \\
\midrule
\multirow{5}{*}{Font Size}
& 8 pt & 0.7222 \\
& 10 pt & 0.7677 \\
& 12 pt & 0.7626 \\
& 14 pt & \underline{0.7727} \\
& 16 pt & \textbf{0.7727} \\
\midrule
\multirow{3}{*}{Text Width}
& 2.0 in & \textbf{0.7929} \\
& 4.0 in & \underline{0.7576} \\
& 6.0 in & 0.7475 \\
\bottomrule
\end{tabular}
\caption{GPQA Diamond accuracy of GPT-5.1 under different T-OR renderer ablations.}
\label{tab:ablation_exp}
\end{table}

%% file: tables/extreme_compression.tex
\begin{table}[!t]
\small
\centering
\begin{tabular}{lccc}
\toprule
\textbf{Setting} & \textbf{\begin{tabular}[c]{@{}c@{}}Reasoning \\ Tokens\end{tabular}} & \textbf{\begin{tabular}[c]{@{}c@{}}Token \\ Reduction\end{tabular}} & \textbf{Acc.} \\
\midrule
No reasoning & 0.0 & -100\% & 0.6890 \\
\midrule
Text reasoning & 95.3 & -0\% & 0.7323 \\
\midrule
\multirow{9}{*}{\textbf{T-OR}}
& 1.2 & -98.75\% & 0.7008 \\
& 2.4 & -97.50\% & 0.7283 \\
& 4.8 & -95.00\% & 0.7638 \\
& 7.2 & -92.50\% & \textbf{0.7992} \\
& 19.1 & -80.00\% & 0.7677 \\
& 38.2 & -60.00\% & 0.7520 \\
& 57.4 & -40.00\% & \underline{0.7835} \\
& 76.4 & -20.00\% & 0.7402 \\
& 95.6 & -0.00\% & 0.7362 \\
\bottomrule
\end{tabular}
\caption{Accuracy on AquaRat under extreme reasoning-token compression using T-OR with Gemini 2.5 Flash.}
\vspace{-1em}
\label{tab:extreme-compression}
\end{table}

%% file: tables/render.tex
\begin{table}[!t]
\small
\centering
\resizebox{\columnwidth}{!}{%
\begin{tabular}{@{}lccc@{}}
\toprule
\textbf{Model} & \textbf{Pillow} & \textbf{Matplotlib} & \textbf{XeLaTeX} \\
\midrule
Qwen3-VL-235B     & 0.6970 & 0.6970 & 0.7323 \\
Kimi K2.5         & 0.7576 & 0.7273 & 0.7626 \\
Gemini 2.5 Flash  & \textbf{0.8131} & \textbf{0.8182} & \underline{0.7828} \\
Claude Sonnet 4.5 & \underline{0.7929} & \underline{0.7727} & \textbf{0.8131} \\
\bottomrule
\end{tabular}
}
\caption{Accuracy on GPQA Diamond using T-OR with different renderers. Pillow, Matplotlib, and XeLaTeX denote the rendering backends used to convert the same textual rationale into typographical rationale.}
\label{tab:render_exp}
\vspace{-1em}
\end{table}

%% file: tables/efficient_text_reasoning.tex
\begin{table}[!t]
\small
\centering
\begin{tabular}{lcc}
\toprule
\textbf{Setting} & \textbf{Reasoning Tokens} & \textbf{Acc.} \\
\midrule
No reasoning & 0.0 & 0.6890 \\
\midrule
Text reasoning & 95.0 & 0.7323 \\
\midrule
\multirow{5}{*}{LLMLingua-2}
& 17.7 & 0.6890 \\
& 37.7 & 0.6890 \\
& 57.6 & 0.6929 \\
& 76.8 & 0.6929 \\
\midrule
\multirow{5}{*}{T-OR}
& 19.1 & \underline{0.7677} \\
& 38.2 & 0.7520 \\
& 57.4 & \textbf{0.7835} \\
& 76.4 & 0.7402 \\
\bottomrule
\end{tabular}
\caption{AquaRat results with LLMLingua-2 and T-OR.}
\label{tab:efficient-text-reasoning}
\end{table}

%% file: tables/free_reasoning.tex
\begin{table}[!t]
\small
\centering
\begin{tabular}{l|c|c}
\toprule
\textbf{Setting} & \textbf{Token Reduction} & \textbf{Acc.} \\
\midrule
No reasoning & -100\% & 0.4646 \\
\cmidrule(lr){1-3}
Free reasoning & -0\% & \underline{0.6869} \\ 
\cmidrule(lr){1-3}
\multirow{5}{*}{T-OR} & -80\% & 0.6162 \\
& -60\% & 0.6616 \\
& -40\% & 0.6768 \\
& -20\% & \underline{0.6869} \\
& -0\% & \textbf{0.6919} \\ 
\bottomrule
\end{tabular}
\caption{Accuracy on GPQA Diamond using T-OR with rationales generated by GPT-5.1. We compare no reasoning, free reasoning under different budgets.}
\label{tab:free-exp}
\vspace{-1em}
\end{table}

%% file: 5_conclusion.tex
\section{Conclusion and Future Work}
In this study, we propose optical reasoning, rethinking images as the sole reasoning medium, and identify two core potentials. First, rendering rationales into images can preserve reasoning performance while substantially improving efficiency. Second, images provide a unified visual canvas that naturally integrates text, graphical elements, and spatial layouts. The typographic-based variant validates the former potential by using an optimal layout token strategy to maximize information density. The graphical-based variant targets the latter potential by organizing rationales into step-aligned compositions. Empirical results across diverse benchmarks and MLLMs show that: images can serve as an effective and efficient reasoning medium and exhibit unique capabilities for structuring interleaved-modal rationales. A key future direction is to mitigate graphical hallucinations in expressive rationale generation, thereby further unlocking the potential of images as an independent reasoning medium.
\section*{Limitations}
This work has two limitations. 1) Model-dependent perception. Optical reasoning may behave differently across MLLMs. Its effectiveness can be affected by model-specific sensitivity to resolution, layout density, rendering style, and visual-token budgets. Future work may develop model-adaptive rendering strategies to improve robustness across different models. 2) Reliability of graphical rationales. Generated schematics may contain graphical inaccuracies. While graphical-based optical reasoning offers stronger visual expressiveness, its reliability can be improved through end-to-end fine-tuning or reinforcement learning with feedback on visual correctness and answer accuracy.
\section*{Ethics Statement}
This work uses publicly available reasoning benchmarks and does not collect new human-subject data. The datasets are used only for research evaluation and do not contain private user information to the best of our knowledge. The models, tools, and scientific artifacts used in this work are publicly accessible or available through official APIs, and are used in accordance with their intended purposes. The generated rationale images are only intermediate representations for benchmark evaluation rather than human-facing outputs or deployment decisions. Therefore, we believe this work poses minimal ethical risk.

%% file: 6_appendix.tex
\section{Case Study}
\label{subsec:case_study}

\begin{figure*}[t!]
\centering
  \includegraphics[width=1.0\textwidth]{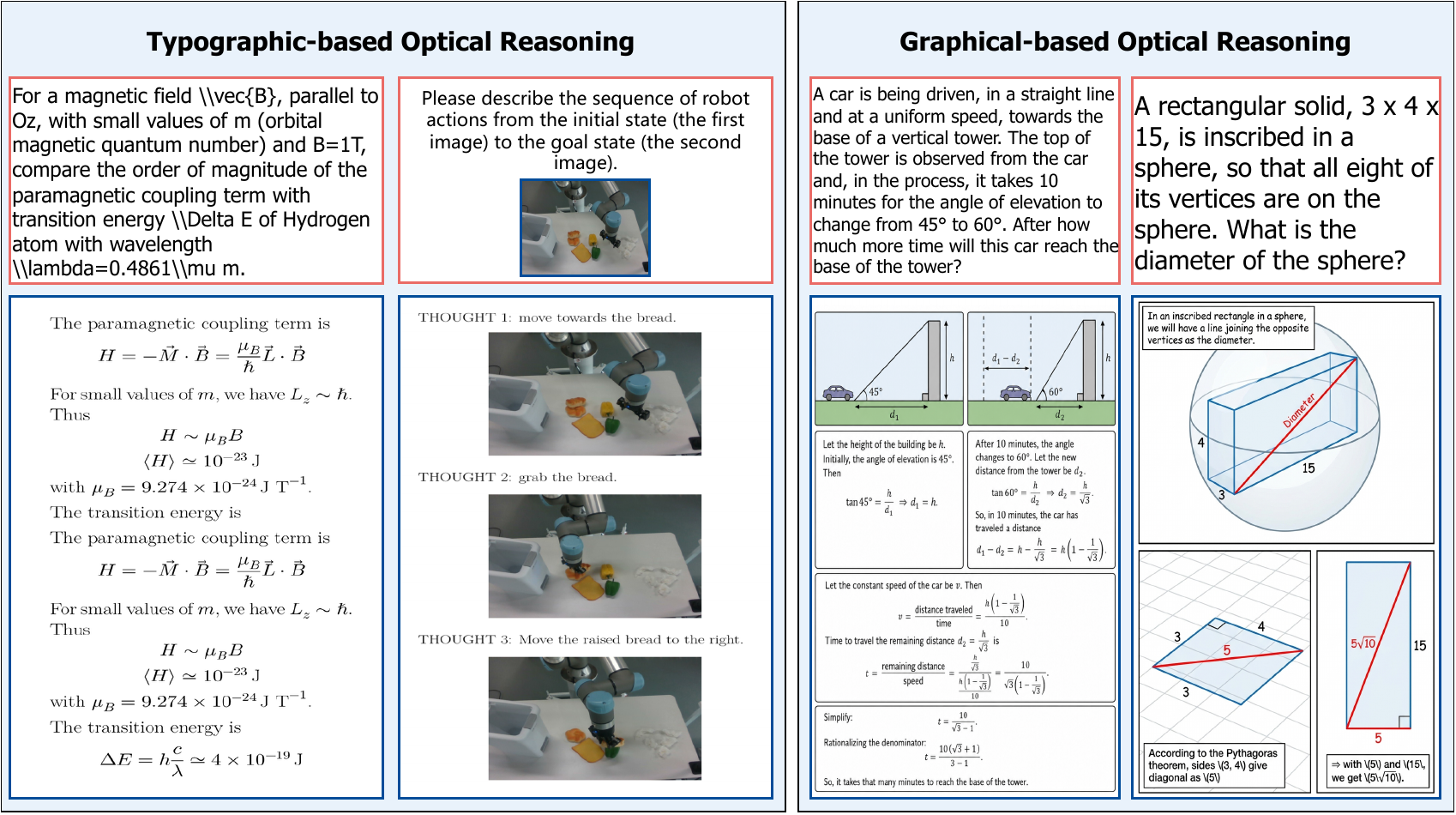}
  \caption{Case study comparing T-OR and G-OR across text-only and multimodal reasoning tasks. T-OR preserves the original rationale in a dense typographic layout, while G-OR reorganizes the rationale into step-aligned graphical panels.}
  \label{fig:case_study_1}
\end{figure*}

To qualitatively illustrate why optical reasoning works, we analyzed the rationales from different modes in Figure~\ref{fig:case_study_1}. 
First, both T-OR and G-OR effectively integrate text, graphical elements, and spatial layouts within a unified visual canvas. T-OR faithfully preserves the original rationale. For example, the physics case retains key equations and derivations, while the robot case aligns action descriptions with corresponding visual thoughts. G-OR goes one step further by adopting a more flexible spatial organization, using panels and diagrammatic decomposition to separate different reasoning stages. As shown in the tower and geometry cases, G-OR explicitly visualizes intermediate variables and spatial relations, enabling the rationale to be conveyed through multimodal illustrations. These observations demonstrate that images, as an expressive reasoning medium, can naturally support interleaved-modal CoT and further strengthen reasoning by exploiting spatial relations within a unified visual canvas.
However, we also revealed the limitations of G-OR. 

\begin{figure*}[t!]
\centering
  \includegraphics[width=1.0\textwidth]{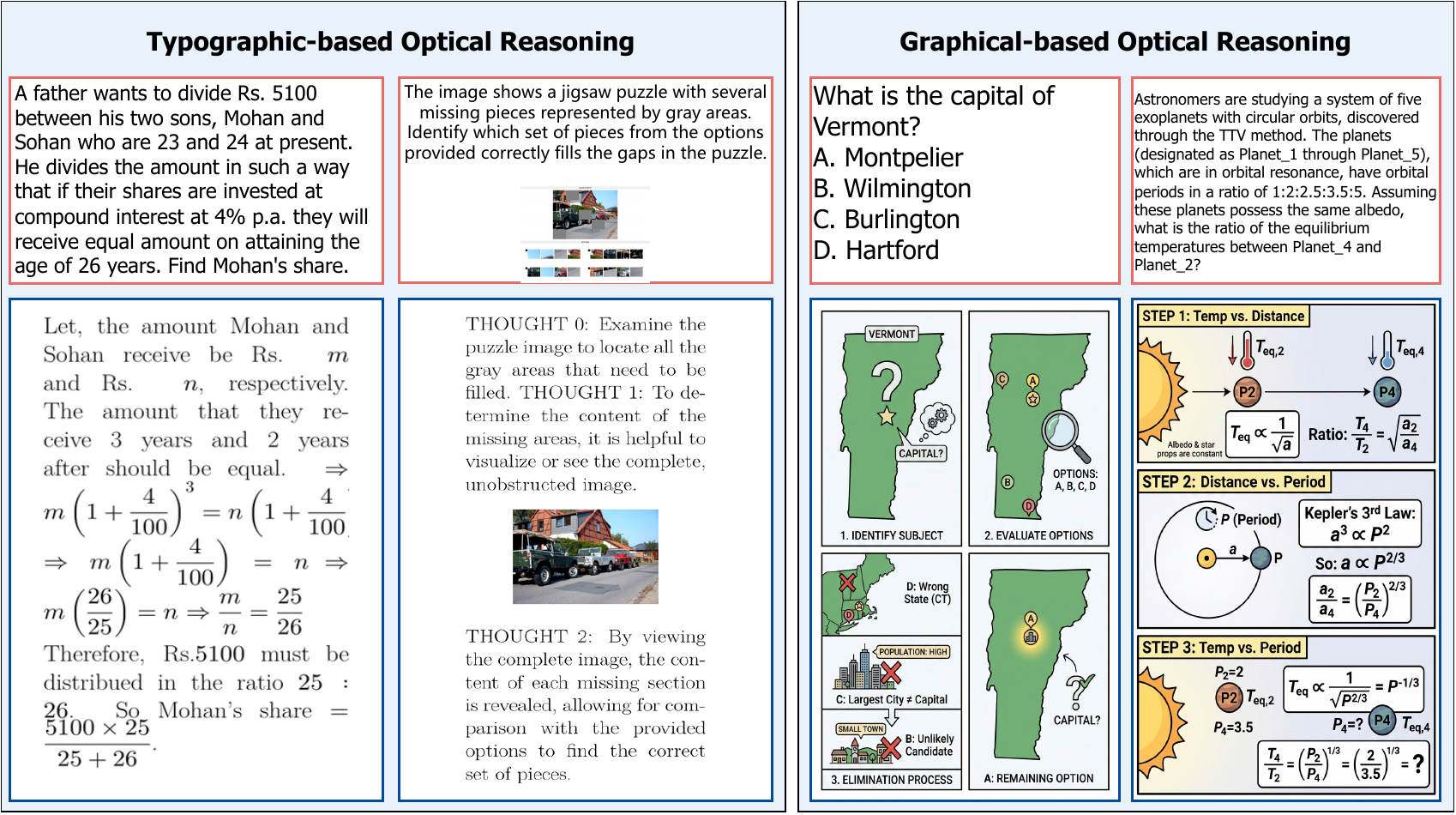}
  \caption{Additional illustrations of T-OR and G-OR across text-only and multimodal reasoning tasks.}
  \label{fig:case_study_2}
\end{figure*}

\begin{figure*}[t!]
\centering
  \includegraphics[width=1.0\textwidth]{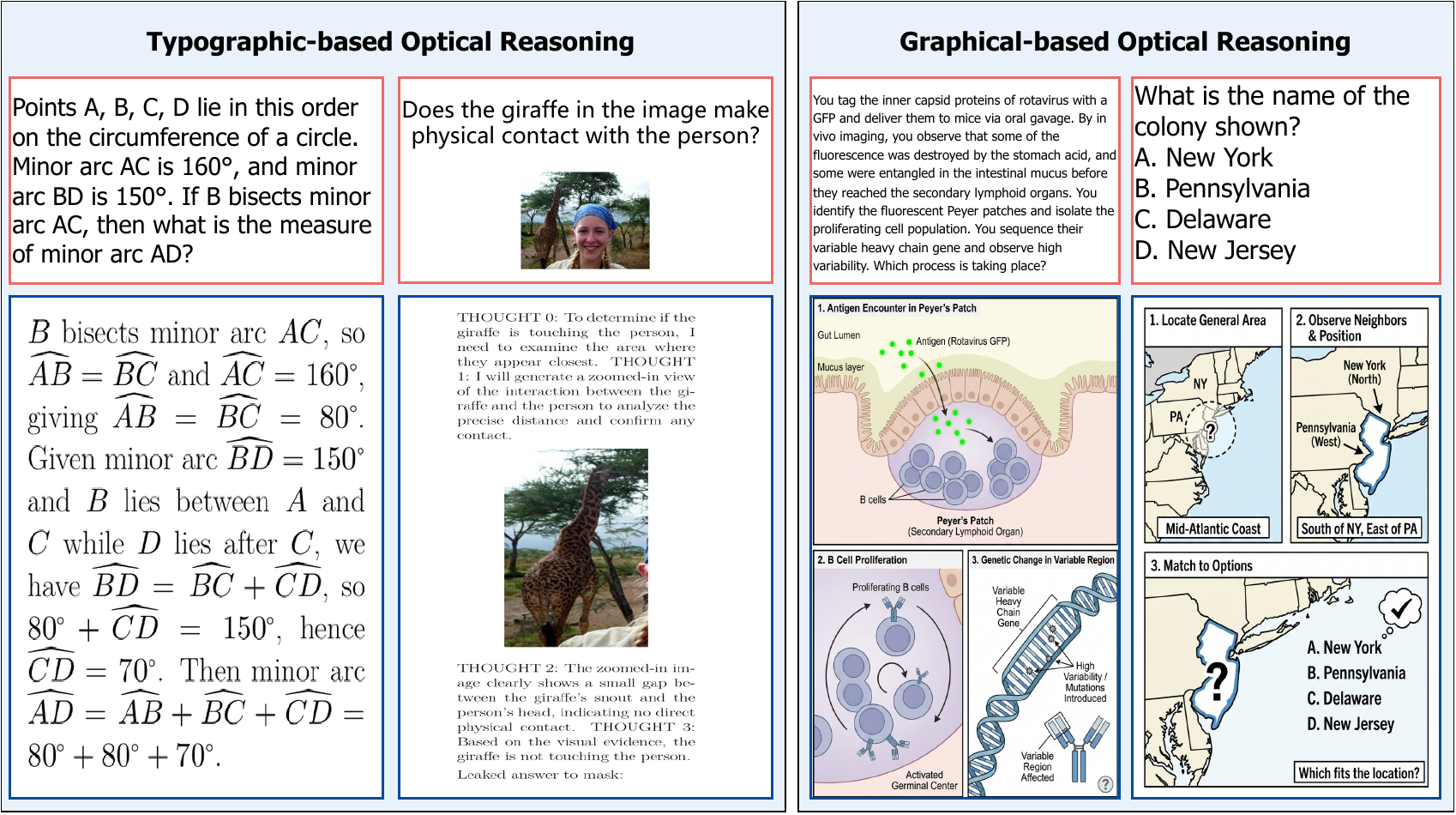}
  \caption{Additional illustrations of T-OR and G-OR across text-only and multimodal reasoning tasks.}
  \label{fig:case_study_3}
\end{figure*}

\begin{figure*}[t!]
\centering
  \includegraphics[width=1.0\textwidth]{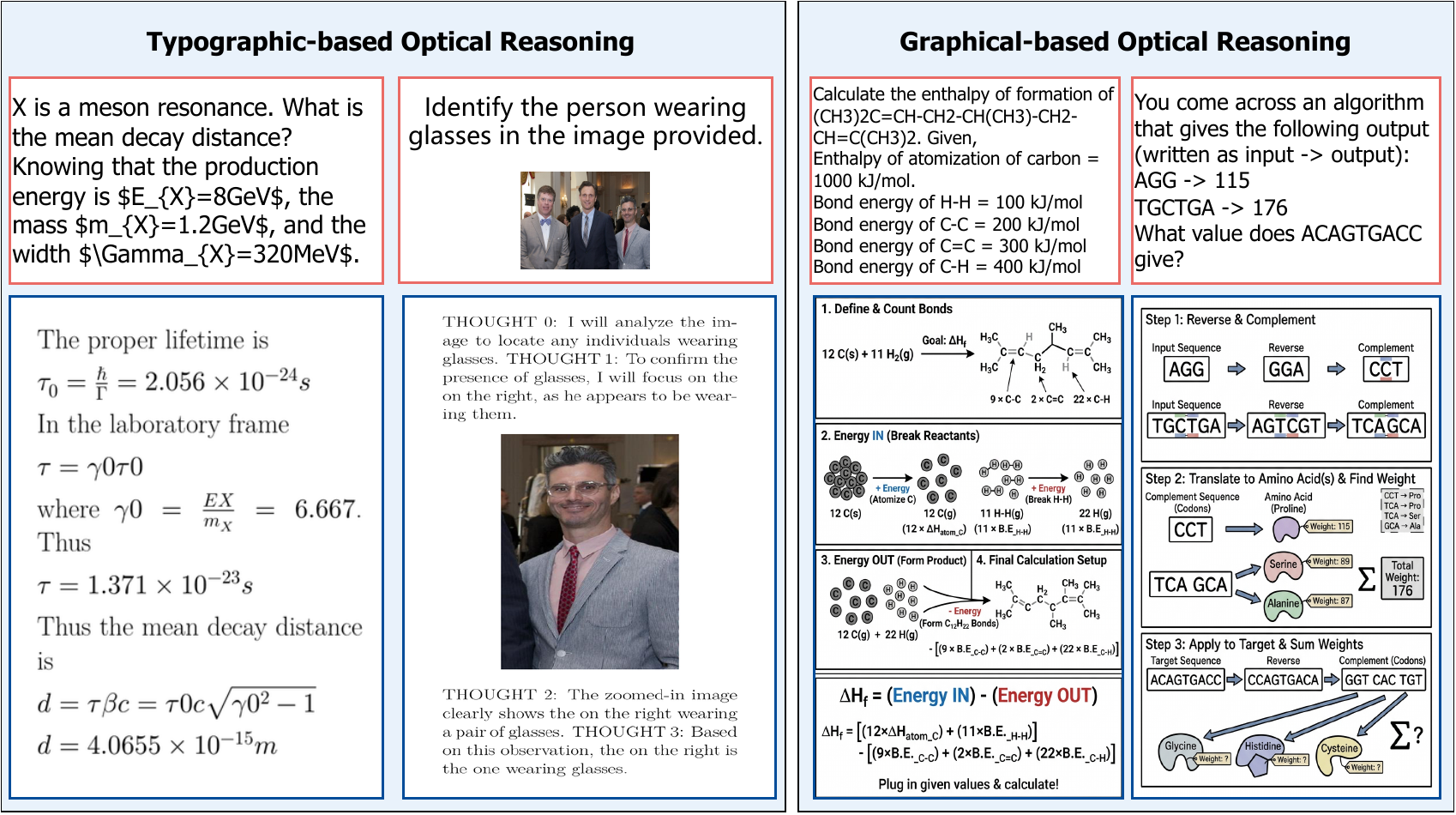}
  \caption{Additional illustrations of T-OR and G-OR across text-only and multimodal reasoning tasks.}
  \label{fig:case_study_4}
\end{figure*}

Although graphical elements can vividly express complex concepts, the generated schematics are not always accurate. For example, in the geometric case, the red segment is intended to indicate the key diagonal relation, but its placement deviates from the exact geometric constraint. This suggests that G-OR provides stronger visual expressiveness, but may also introduce graphical hallucination as a new failure mode.
More case studies are provided in Figures~\ref{fig:case_study_2}, \ref{fig:case_study_3} and \ref{fig:case_study_4}.

\section{Prompt Templates}
This section provides the prompt templates used in our evaluation. 
\subsection{Evaluation Prompts}
\label{appendix:evaluation-prompts}
For answer prediction, all evaluated models are instructed to output only the final answer in a boxed format. Table~\ref{tab:evaluation-prompts} first lists the baseline prompt settings: no reasoning receives only the problem text, text reasoning receives the problem followed by the rationale, and free reasoning asks the model to solve the problem step by step. For optical reasoning, T-OR takes the problem text together with the rendered typographic rationale image, while G-OR takes the problem text together with the generated graphical rationale image. For datasets with fixed answer choices, the system prompt is specialized to require exactly one boxed choice label.
\begin{table*}[t]
\centering
\small
\setlength{\tabcolsep}{6pt}
\renewcommand{\arraystretch}{1.12}
\begin{tabular}{p{0.20\textwidth} p{0.74\textwidth}}
\toprule
Setting & Prompt template \\
\midrule
No reasoning &
\textbf{System:} You are an expert problem solver. Your only task is to provide the final answer in \(\backslash\)boxed\{ANSWER\} format. Do not show your work, intermediate steps, or reasoning.\newline
\textbf{User:} \{Problem\}. \\
\midrule
Text reasoning &
\textbf{System:} You are an expert problem solver. Your only task is to provide the final answer in \(\backslash\)boxed\{ANSWER\} format. Do not show your work, intermediate steps, or reasoning.\newline
\textbf{User:} \{Problem\}\newline\newline\{rationale\}. \\
\midrule
Free reasoning &
\textbf{System:} You are an expert problem solver. Solve the problem step by step. At the end, provide your final answer in the format \(\backslash\)boxed\{ANSWER\}.\newline
\textbf{User:} \{Problem\}. \\
\midrule
T-OR &
\textbf{System:} You are an expert problem solver. Your only task is to provide the final answer in \(\backslash\)boxed\{ANSWER\} format. Do not show your work, intermediate steps, or reasoning.\newline
\textbf{User:} \{Problem\}. The rendered typographic rationale image is provided as a separate visual input. \\
\midrule
G-OR &
\textbf{System:} You are an expert problem solver. Your only task is to provide the final answer in \(\backslash\)boxed\{ANSWER\} format. Do not show your work, intermediate steps, or reasoning.\newline
\textbf{User:} \{Problem\}. The generated graphical rationale image is provided as a separate visual input. \\
\end{tabular}
\caption{Evaluation prompt templates for the main task settings.}
\label{tab:evaluation-prompts}
\end{table*}
\subsection{Graphical Rationale Generation Prompt}
\label{appendix:t2i}
For G-OR, we use Nano Banana 2 to transform the problem and answer-masked rationale into a compact rationale image. The prompt in Table~\ref{tab:t2i-prompt} operationalizes G-OR by converting the answer-masked rationale into a step-aligned multi-panel visual canvas, where textual reasoning anchors, graphical elements, and spatial layouts jointly carry the reasoning process.
\begin{table*}[t]
\centering
\small
\setlength{\tabcolsep}{6pt}
\renewcommand{\arraystretch}{1.08}
\begin{tabular}{p{0.96\textwidth}}
\toprule
\textbf{Prompt template} \\
\midrule
You are an expert educational illustrator.\newline
\textbf{Task:} Create a compact, step-by-step comic-style illustration that explains how to solve a problem.\newline
\textbf{Input:} Question: \{Problem\}. Solution: \{Answer-masked rationale\}.\newline
\textbf{Strict requirements:} The comic must follow the solution steps, but must not reveal the final answer explicitly. Preserve the reasoning text, transformations, and intermediate expressions from the provided solution visibly in the image. Include only visuals that directly help explain the solving process. Do not include irrelevant people, characters, decorations, or background scenery. Break the solution into 2--4 clear logical steps, with one panel per step.\newline
\textbf{Layout and compactness:} Use a tight multi-panel layout, either 2--4 panels in a single horizontal row or a compact 2-by-2 grid if needed. Keep panel spacing narrow, outer margins small, and content dense enough to fill most of the canvas. Keep each panel simple, information-rich, consistently sized, and easy to scan.\newline
\textbf{Visual style:} Clean educational comic style, simple shapes, clear labels, high contrast, plain white background, minimal clutter, crisp linework, and readable mathematical notation.\newline
\textbf{Output:} A single compact multi-panel comic illustration. Each panel should correspond to one logical step in the solution. Preserve the reasoning text visibly in the image, but do not include or complete the final answer. \\
\bottomrule
\end{tabular}
\caption{Prompt template for generating graphical rationales in G-OR.}
\label{tab:t2i-prompt}
\end{table*}
\subsection{LLM Judge Prompt}
\label{appendix:llm_judge}
We first apply rule-based answer extraction and matching. For predictions that cannot be resolved reliably by the rule-based matcher, we use an LLM judge as a fallback. The judge prompt in Table~\ref{tab:llm-judge-prompt} is constrained to compare only the final prediction against the gold answer and to return a structured correctness verdict.
\begin{table*}[t]
\centering
\small
\setlength{\tabcolsep}{6pt}
\renewcommand{\arraystretch}{1.12}
\begin{tabular}{p{0.96\textwidth}}
\toprule
\textbf{LLM judge template} \\
\midrule
\textbf{System:} You are an exact grader for final answers. Decide whether the student's final answer is equivalent to the gold answer. Focus only on final-answer equivalence. Treat semantically equivalent, numerically equivalent, or format-equivalent answers as correct when they express the same answer. If the student's answer is ambiguous, incomplete, or includes multiple conflicting final answers, mark it incorrect. Do not reward correct reasoning if the final answer is wrong. Return JSON only: \{"verdict":"CORRECT|INCORRECT"\}.\newline
\textbf{User:} Problem: \{Problem\}. Gold answer: \{Gold answer\}. Student answer: \{Model answer\}. \\
\bottomrule
\end{tabular}
\caption{LLM judge prompt template for unresolved answer matching cases.}
\label{tab:llm-judge-prompt}
\end{table*}
\section{Patch Token Mapping}
\label{appendix:patch-token-mapping}
Following the setting of CodeOCR~\citep{shi_codeocr_2026}, we uniformly apply a Qwen3-VL-style patch mapping~\citep{bai2025qwen3vltechnicalreport} across all models to estimate visual reasoning tokens. Concretely, for a rendered reasoning image with height \(H\) and width \(W\), we estimate its visual token count as
\begin{equation}
    N = \left\lceil \frac{H}{32} \right\rceil
        \left\lceil \frac{W}{32} \right\rceil .
\end{equation}
Closed-source models~\citep{singh2026openaigpt5card, comanici2025gemini25pushingfrontier} do not publicly disclose their exact internal visual tokenization rules. We therefore use the same patch-based estimator for all models to keep token accounting comparable. When generating T-OR variants, we resize the rendered rationale image to match target token budgets corresponding to reduction ratios from \(-80\%\) to \(0\%\) relative to the text-reasoning token count. This procedure enables controlled comparisons between text reasoning and visual reasoning under matched or compressed reasoning budgets.
\begin{table*}[t]
\centering
\small
\setlength{\tabcolsep}{5pt}
\renewcommand{\arraystretch}{1.12}
\begin{tabular}{p{0.18\textwidth} p{0.48\textwidth} p{0.10\textwidth} p{0.18\textwidth}}
\toprule
Benchmark & Task & Size & Metric \\
\midrule
AquaRat~\citep{DBLP:journals/corr/LingYDB17} &
Multiple-choice algebra and quantitative reasoning problems with five answer options. &
254 &
Accuracy / MAG \\
\midrule
Gsm8k~\citep{cobbe2021gsm8k} &
Open-ended grade-school math word problems requiring multi-step arithmetic reasoning. &
1{,}319 &
Accuracy / MAG \\
\midrule
GPQA Diamond~\citep{rein2023gpqagraduatelevelgoogleproofqa} &
Graduate-level multiple-choice science questions spanning physics, chemistry, and biology. &
198 &
Accuracy / MAG \\
\midrule
ScienceQA~\citep{lu2022learnexplainmultimodalreasoning} &
Multimodal science question answering with image inputs for visually grounded examples. &
1{,}836 &
Accuracy / MAG \\
\midrule
Zebra-CoT~\citep{li2025zebracotdatasetinterleavedvision} &
Interleaved text-image visual reasoning problems containing question images and intermediate visual rationale images. &
300 &
Accuracy / MAG \\
\bottomrule
\end{tabular}
\caption{Benchmark details for the main evaluation. Dataset sizes correspond to the processed JSONL files used in our experiments.}
\label{tab:benchmark-details}
\end{table*}
\section{Benchmark Details}
\label{appendix:bench}
Table~\ref{tab:benchmark-details} summarizes benchmark categories, task formats, sizes, and evaluation metrics used in our experiments. We report accuracy for task performance and MAG for token efficiency, following Eq.~\ref{eq:mag}.
For Zebra-CoT~\citep{li2025zebracotdatasetinterleavedvision}, we construct the evaluation split reproducibly by applying reservoir sampling to each source subset with a fixed random seed of 42. The final evaluation set contains 300 examples from 15 sampled subsets: 2D Visual Reasoning (Visual Jigsaw, Visual Search), 3D Visual Reasoning (Embodied CoT, Robot Planning), Scientific Reasoning (Chemistry, Competitive Programming, Geometry, Physics), and Visual Logic \& Strategic Games (Checkers, Chess, Ciphers, Connect Four, Maze, RPM, Tetris).

